\documentclass[sigconf]{acmart}
\usepackage{framed}
\usepackage[table]{xcolor}
\usepackage{booktabs}
\usepackage{graphicx}
\usepackage{pifont}
\usepackage{fancyvrb}

\definecolor{lightgraybg}{gray}{0.95}
\newcommand{\reviewhead}[1]{%
  \noindent
  \fcolorbox{black}{black}{%
    \parbox{\dimexpr\columnwidth-2\fboxsep-2\fboxrule\relax}{%
      \textcolor{white}{\textbf{#1}}%
    }%
  }%
}
\AtBeginDocument{%
  }

\usepackage{gradient-text}
\usepackage{xcolor}
\usepackage{svg}
\usepackage{pifont}
\usepackage{multibib}
\newcites{supp}{Supplementary References}
\definecolor{headergray}{RGB}{240,240,240}
\definecolor{finrow}{RGB}{230,245,255}
\definecolor{highlight}{RGB}{255,250,205}

\setcopyright{none} 
\renewcommand\footnotetextcopyrightpermission[1]{} 
\copyrightyear{2018}
\acmYear{2018}
\acmDOI{XXXXXXX.XXXXXXX}
\acmConference[], {,
  }{}
\acmISBN{978-1-4503-XXXX-X/2018/06}

\usepackage{xcolor}
\usepackage{pifont}
\usepackage{colortbl} 
\usepackage{threeparttable}
\usepackage{svg}
\usepackage{xspace}
\usepackage{gradient-text}
\newcommand{\ficriticaled}{%
  \textsc{\gradientRGB{FinCriticalED}{50,180,140}{0,0,185}}\xspace%
}
\usepackage{booktabs}
\usepackage{colortbl}
\usepackage{xcolor}
\usepackage{pifont}
\newcommand{\cmark}{\textcolor{green!60!black}{\ding{52}}}
\newcommand{\xmark}{\textcolor{red}{\ding{56}}}
\definecolor{lightblue}{RGB}{219,234,254}

\usepackage{subcaption}
\usepackage{booktabs}
\usepackage{makecell}
\usepackage{tabularx}

\begin{document}

\title{\ficriticaled: A Visual Benchmark for Financial Fact-Level OCR}


\author{Yueru He}
\affiliation{%
  \institution{\normalsize Columbia University}
  \country{\normalsize USA}
}
\email{yh3507@columbia.edu}

\author{Yupeng Cao}
\affiliation{%
  \institution{\normalsize Stevens Institute of Technology}
  \country{\normalsize USA}
}

\author{Yan Wang}
\affiliation{%
  \institution{\normalsize The Fin AI}
  \country{\normalsize USA}
}

\author{Lingfei Qian}
\affiliation{%
  \institution{\normalsize The Fin AI}
  \country{\normalsize USA}
}

\author{Shuyao Wang}
\affiliation{%
  \institution{\normalsize The Fin AI}
  \country{\normalsize USA}
}

\author{Yi Han}
\affiliation{%
  \institution{\normalsize Georgia Institute of Technology}
  \country{\normalsize USA}
}

\author{Haohang Li}
\affiliation{%
  \institution{\normalsize Stevens Institute of Technology}
  \country{\normalsize USA}}

\author{Ruoyu Xiang}
\affiliation{%
  \institution{\normalsize New York University}
  \country{\normalsize USA}
}

\author{Fan Zhang}
\affiliation{%
  \institution{\normalsize The University of Tokyo \& MBZUAI}
  \country{\normalsize Japan \& UAE}
}

\author{Zhuohan Xie}
\affiliation{%
  \institution{\normalsize MBZUAI}
  \country{\normalsize UAE}
}

\author{Mingquan Lin}
\affiliation{%
  \institution{\normalsize University of Minnesota}
  \country{\normalsize USA}
}

\author{Prayag Tiwari}
\affiliation{%
  \institution{\normalsize Halmstad University}
  \country{\normalsize USA}
}

\author{Xueqing Peng}
\authornote{Corresponding author.}
\affiliation{%
  \institution{\normalsize The Fin AI}
  \country{\normalsize USA}
}
\email{xueqing.peng2024@gmail.com}

\author{Jimin Huang}
\affiliation{%
  \institution{\normalsize The Fin AI}
  \country{\normalsize USA}
}

\author{Guojun Xiong}
\affiliation{%
  \institution{\normalsize Harvard University}
  \country{\normalsize USA}
}

\author{Sophia Ananiadou}
\affiliation{%
  \institution{\normalsize University of Manchester}
  \country{\normalsize UK}
}

\renewcommand{\shortauthors}{He et al.}

\begin{abstract}

Recent progress in multimodal large language models (MLLMs) has substantially improved document understanding, yet strong optical character recognition (OCR) performance on surface metrics does not guarantee faithful preservation of decision-critical evidence. This limitation is especially consequential in financial documents, where small visual errors can induce discrete shifts in meaning. To study this gap, we introduce \ficriticaled~(\textbf{Fin}ancial \textbf{Critical} \textbf{E}rror \textbf{D}etection), a fact-centric visual benchmark for evaluating whether OCR and vision-language systems preserve financially critical evidence beyond lexical similarity. \ficriticaled contains 859 real-world financial document pages with 9,481 expert-annotated facts spanning five critical field types: \textit{numeric}, \textit{temporal}, \textit{monetary unit}, \textit{reporting entity}, and \textit{financial concept}. We formulate the task as structured OCR with fact-level verification, and develop a Deterministic-Rule-Guided LLM-as-Judge protocol to assess whether model outputs preserve annotated facts in context. We benchmark 13 systems spanning OCR pipelines, specialized OCR VLMs, open-source MLLMs, and proprietary MLLMs. Results reveal a clear gap between lexical accuracy and factual reliability, with numerical values and monetary units emerging as the most vulnerable fact types, and critical errors concentrating in visually complex, mixed-layout documents with distinct failure patterns across model families. Overall, \ficriticaled provides a rigorous benchmark for trustworthy financial OCR and a practical testbed for evidence fidelity in high-stakes multimodal document understanding. The dataset is available at: {\color{blue}\url{https://the-finai.github.io/FinCriticalED/}}.
\end{abstract}



\received{-}
\received[revised]{-}
\received[accepted]{-}


\begin{teaserfigure}
\centering
\setlength{\abovecaptionskip}{0pt}
\setlength{\belowcaptionskip}{0pt}
\includegraphics[width=0.96\linewidth]{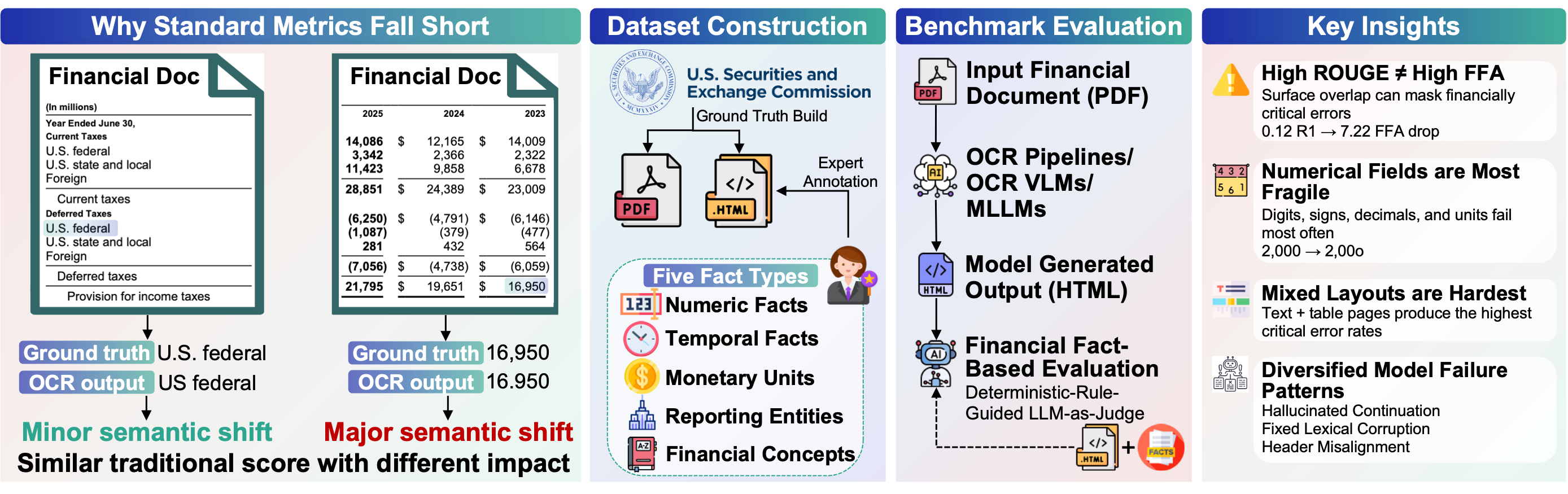}
\caption{
Overview of \ficriticaled. Left: standard OCR metrics fail to capture financially critical semantic errors. Middle-left: construction of a fact-centric dataset with five types of annotated financial facts. Middle-right: evaluation pipeline using OCR/VLM/MLLM outputs with Deterministic-Rule-Guided LLM-as-Judge. Right: Key insights from experiments on 13 OCR systems and MLLMs.
}
\label{fig:overall}
\end{teaserfigure}

\maketitle

\section{Introduction}

Recent progress in multimodal large language models (MLLMs) has made document understanding appear increasingly successful~\citep{fu2025ocrbenchv2improvedbenchmark,ouyang2024omnidocbenchbenchmarkingdiversepdf}, yet even large models may capture the gist of a page while missing the small visual cues that determine whether a statement is true. When truth and decision-making depend on such cues, the question is no longer how well a model summarizes a document, but whether it understands it in any factually meaningful sense. Financial optical character recognition (OCR) makes this boundary unusually sharp. Decimal points, negative markers, unit scales, currency symbols, and row-column alignments can each induce discrete shifts in meaning~\citep{Watson_2020,ouyang2024omnidocbenchbenchmarkingdiversepdf}. These cases expose a central fault line in current MLLMs: whether they preserve the local evidence on which document truth depends, or whether much of their apparent success still rests on semantic approximation, the distinction directly determines whether a transcribed document can be trusted. This question matters not only for understanding the limits of MLLMs, but also for high-stakes financial analysis, where even a minor extraction error can cascade into large reliability gap and materially false interpretations.

Existing benchmarks remain poorly aligned with this gap, since outputs can stay lexically close to the source while becoming factually wrong because of visually subtle but decisive cues. General OCR benchmarks~\cite{wang2025finauditing, Liu_2024, 
qian2025fino1,fu2025ocrbenchv2improvedbenchmark, yang2024ccocrcomprehensivechallengingocr} primarily assess transcription fidelity and are therefore poorly suited to cases where small local deviations induce asymmetrically large shifts in meaning. Broader document understanding benchmarks~\citep{mathew2020docvqa,mathew2021infographicvqa,vanlandeghem2023documentunderstandingdatasetevaluation} emphasize page-level reasoning, layout understanding, or open-ended question answering, but typically do not isolate whether models preserve the visually grounded local evidence that determines factual correctness. Even finance-oriented benchmarks~\cite{peng2025multifinbenbenchmarkinglargelanguage,masry2024longfinmultimodaldocumentunderstanding}, while domain-relevant, are generally not constructed around this failure mode. At the dataset level, they rarely center financially decisive visual anchors as the unit of evaluation, and at the metric level, they seldom distinguish harmless lexical noise from materially false extraction.

To bridge this gap, we introduce \ficriticaled (\textbf{Fin}ancial \textbf{Critical} \textbf{E}rror \textbf{D}etection), a visual benchmark for evaluating whether OCR and vision-language systems faithfully preserve financially critical evidence. The dataset contains 9,481 expert-annotated financial facts from 859 complex pages across five categories—\textit{numeric}, \textit{temporal}, \textit{monetary unit}, \textit{reporting entity}, and \textit{financial concept}, with 150+ expert annotation hours and an inter-annotator agreement of 0.88. We formulate the benchmark as a structured OCR task with fact-level verification. Given a financial document page, a system produces an HTML-formatted OCR output that preserves both textual content and layout structure, and the benchmark tests whether this output preserves the annotated financial facts without factual distortion. Because financial facts do not map cleanly onto surface forms, string overlap alone is insufficient: the same fact may admit multiple valid expressions (e.g., “\$1.2B” vs. “1,200 million”), while a lexically minor deviation may already make the extracted fact false. We therefore evaluate \ficriticaled at the fact level, using a Deterministic-Rule-Guided LLM-as-Judge verification paradigm to verify whether each annotated fact is preserved in context, with reliability further examined against human experts and reaching 95\% alignment after three rounds of prompt adjustment and output post-processing.

Our benchmarking of 13 state-of-the-art systems reveals a clear gap between surface-level OCR quality and financially faithful document understanding. While several models achieve near-saturated lexical scores (up to $98.84\%$ ROUGE-1), fact-level performance varies substantially, with overall Financial Fact Accuracy (FFA) ranging from $65.68\%$ to $97.23\%$. Notably, small lexical differences can conceal large disparities in financial reliability: a gap of only $0.12$ ROUGE-1 points can correspond to a $7.22$ drop in FFA, highlighting the limitations of overlap-based metrics.
Numerical values and monetary units are consistently the most vulnerable, while temporal and conceptual facts, although relatively better, remain insufficiently reliable for high-stakes use. Critical errors concentrate in visually complex and mixed-layout documents, and failure patterns differ across model families: OCR-specialized systems exhibit systematic recognition artifacts, whereas generative MLLMs are prone to hallucinated continuations and context-driven distortions.
Together, these findings indicate that current models capture surface content well but fail to reliably preserve the localized, visually grounded evidence required for financially correct interpretation.

\section{Related Work}




\begin{table}[bp]
\setlength{\abovecaptionskip}{3pt}
\setlength{\belowcaptionskip}{0pt}
\caption{Comparison of multimodal OCR and financial benchmarks. \cmark indicates inclusion and \xmark exclusion of a property.}
\label{tab:related_work}
\centering
\small
\resizebox{\linewidth}{!}{ %
\begin{tabular}{lccccc}
\rowcolor{headergray}
\toprule
\textbf{Benchmark} & \textbf{Financial} & \textbf{OCR} & \textbf{Page-Level} & \textbf{Asymmetric Error} & \textbf{Critical-Field} \\
\rowcolor{headergray}
 & \textbf{Centric} & \textbf{Included} & \textbf{Reasoning} & \textbf{Sensitivity} & \textbf{Aware} \\
\midrule

OmniDocBench~\cite{ouyang2024omnidocbenchbenchmarkingdiversepdf} & \xmark & \cmark & \cmark & \xmark & \xmark \\
CC-OCR \cite{yang2024ccocrcomprehensivechallengingocr} & \xmark & \cmark & \xmark & \xmark & \xmark \\
OCRBench / v2 \cite{fu2025ocrbenchv2improvedbenchmark} & \xmark & \cmark & \xmark & \xmark & \xmark \\
OCR-VQA \cite{8978122} & \xmark & \cmark & \xmark & \xmark & \xmark \\
InfographicVQA \cite{mathew2021infographicvqa} & \xmark & \cmark & \cmark & \xmark & \xmark \\
DUDE \cite{vanlandeghem2023documentunderstandingdatasetevaluation} & \xmark & \cmark & \cmark & \xmark & \xmark \\
ChartQA-X \cite{hegde2025chartqaxgeneratingexplanationsvisual} & \xmark & \xmark & \xmark & \xmark & \xmark \\
CONTEXTUAL \cite{wadhawan2024contextualevaluatingcontextsensitivetextrich} & \xmark & \cmark & \xmark & \xmark & \xmark \\
SEED-Bench-2-Plus \cite{li2024seedbench2plusbenchmarkingmultimodallarge} & \xmark & \cmark & \xmark & \xmark & \xmark \\
MMMU\cite{yue2024mmmumassivemultidisciplinemultimodal} / MMMU-Pro\cite{yue-etal-2025-mmmu} & \xmark & \xmark & \xmark & \xmark & \xmark \\
FOX\cite{liu2024focus} & \xmark & \cmark & \xmark & \xmark & \xmark \\
\midrule

\rowcolor{finrow}
MultiFinBen \cite{peng2025multifinbenbenchmarkinglargelanguage} & \cmark & \cmark & \xmark & \xmark & \xmark \\

\rowcolor{finrow}
FinMME \cite{luo2025finmmebenchmarkdatasetfinancial} & \cmark & \xmark & \xmark & \xmark & \xmark \\

\rowcolor{finrow}
MME-Finance \cite{10.1145/3746027.3758230} & \cmark & \xmark & \xmark & \xmark & \xmark \\

\rowcolor{highlight}
\textbf{\textsc{FinCriticalED}} & \textbf{\cmark} & \textbf{\cmark} & \textbf{\cmark} & \textbf{\cmark} & \textbf{\cmark} \\

\bottomrule
\end{tabular}%
}
\end{table} 

\noindent \textbf{OCR Models.}
Document OCR has evolved from modular pipelines to unified multimodal architectures. Traditional frameworks like PaddleOCR~\cite{cui2025paddleocr30technicalreport, cui2025paddleocrvlboostingmultilingualdocument} and MinerU2.5~\cite{niu2025mineru25decoupledvisionlanguagemodel} remain robust for document parsing but often struggle with granular structural fidelity. In contrast, Specialized OCR VLMs (e.g., DeepSeek-OCR~\cite{wei2025deepseek, wei2026deepseek} and GOT-OCR~\cite{wei2023vary, wei2024general}), open-source MLLMs~\citep{google_gemma_3n_E4B_it_2025, qwen3technicalreport, qwen3.5, meta-llama-4-maverick-17b-128e-instruct-fp8} and proprietary models (e.g., GPT~\cite{hurst2024gpt, openai_gpt5}, Gemini~\cite{gemini25pro}, and Claude~\cite{anthropic2026claude46}) leverage visual-to-sequence learning to transcend simple transcription.

\noindent \textbf{General OCR Benchmarks.}
Existing evaluation suites primarily assess either transcription fidelity or high-level reasoning, leaving a critical gap in fact-level evidence preservation. General OCR benchmarks like OCRBench~\cite{Liu_2024, fu2025ocrbenchv2improvedbenchmark}, OmniDocBench~\cite{ouyang2024omnidocbenchbenchmarkingdiversepdf}, and CC-OCR~\cite{yang2024ccocrcomprehensivechallengingocr} evaluate models on character recognition and layout parsing. Conversely, reasoning-centric frameworks like OCR-VQA~\cite{8978122}, ChartQA-X~\cite{hegde2025chartqaxgeneratingexplanationsvisual}, MMMU~\cite{yue2024mmmumassivemultidisciplinemultimodal}, MMMU-Pro~\cite{yue-etal-2025-mmmu}, SEED-Bench-2-Plus~\cite{li2024seedbench2plusbenchmarkingmultimodallarge}, CONTEXTUAL~\cite{wadhawan2024contextualevaluatingcontextsensitivetextrich}, InfographicVQA~\cite{mathew2021infographicvqa}, and FOX~\cite{liu2024focus} emphasize semantic alignment in open-ended QA. Document Understanding Dataset and Evaluation (DUDE)~\cite{vanlandeghem2023documentunderstandingdatasetevaluation} includes both layout and VQA evaluation, but it does not isolate decision-critical fields or measure the asymmetric risk introduced by visually small yet factually consequential errors.

\noindent \textbf{Financial OCR Benchmarks.}
While domain-specific, current financial benchmarks often bypass the core challenges of visual parsing. FinMME~\cite{luo2025finmmebenchmarkdatasetfinancial} and MME-Finance~\cite{10.1145/3746027.3758230} evaluate multimodal QA using pre-cleaned text. While MultiFinBen~\cite{peng2025multifinbenbenchmarkinglargelanguage} includes OCR tasks, it limits evaluation to generic text extraction, failing to weigh the asymmetric importance of critical financial facts. Similarly, LongFin~\cite{masry2024longfinmultimodaldocumentunderstanding} targets long-form representation but does not address visual factual consistency.

Taken together, existing benchmarks stop short of evaluating evidence fidelity under asymmetric risk. \textit{A critical gap remains in critical-field-aware evaluation for OCR in text- and table-dense documents characterized in finance.} \ficriticaled fills this void by centering evaluation on visually-grounded financial anchors where even a single-pixel misinterpretation leads to materially false extraction.

\section{\ficriticaled Benchmark}

To address this gap, we construct \ficriticaled to evaluate OCR systems under high-stakes financial scenarios, shifting the focus from lexical similarity to fact-level evidence preservation.

\subsection{Dataset Curation}

Financial documents are dominated by dense numerical data, structured tables, and specialized terminologies where minor lexical deviations lead to significant factual distortions. Unlike general datasets using surface metrics (e.g., ROUGE, edit distance), \ficriticaled highlights financial critical facts essential for downstream analysis.


\subsubsection{Task Definition}

The task is formulated as a \textit{structured OCR and factual verification} problem. Given an input image $I$, the model generates HTML-formatted OCR output $T = \text{OCR}(I)$ preserving text and layout. To test factual distortion, each ground-truth is paired with annotated facts $F = \{f_1, \dots, f_x\}$, categorized as $F = F_n \cup F_t \cup F_{mu} \cup F_{re} \cup F_{fc}$, representing \textit{numeric} ($F_n$), \textit{temporal} ($F_t$), \textit{monetary unit} ($F_{mu}$), \textit{reporting entity} ($F_{re}$), and \textit{financial concept} ($F_{fc}$) facts (Table~\ref{tab:fact_types}).



\begin{table}[ht]
\setlength{\abovecaptionskip}{3pt}
\setlength{\belowcaptionskip}{0pt}
\centering
\caption{Financial critical fact types in \ficriticaled.}
\label{tab:fact_types}
\scriptsize
\renewcommand{\arraystretch}{1.05}
\begin{tabular}{@{}lp{3.0cm}p{3.0cm}@{}}
\toprule
\textbf{Type} & \textbf{Definition} & \textbf{Examples} \\
\midrule
\textit{Numeric Facts} & Quantitative values, including signed numbers, fractions, and percentages 
& ``2,345''; ``0.37''; ``1/3''; ``-2.3''; ``(10,234)''; ``25.63\%'' \\

\textit{Temporal Facts} & Expressions denoting dates, time periods, or durations 
& ``March 24, 2025''; ``Q2 2025''; ``1 month'' \\

\textit{Monetary Units} & Currency symbols, units, or scale indicators specifying monetary magnitude 
& ``\$''; ``US\$''; ``€''; ``million''; ``thousand'' \\

\textit{Reporting Entities} & Named entities that serve as sources or subjects of reported information 
& ``JPMorgan Chase \& Co.''; ``Alphabet Inc.''; ``CEO'' \\

\textit{Financial Concepts} & Domain-specific financial or accounting concepts 
& ``net income''; ``operating cash flow''; ``EPS''; ``accounts receivable'' \\

\bottomrule
\end{tabular}
\end{table}



\subsubsection{Raw Data Collection}

To ensure visual and structural diversity, we sourced 859 pages from public repositories, including the SEC EDGAR database~\footnote{\url{https://www.sec.gov/search-filings}} and open-source non-profit organization database ~\footnote{\url{https://candid.org/}} (2023–2026). The corpus spans 637 financial legal documents (e.g., M\&A agreements, credit agreements, bond indentures), 142 SEC disclosures (e.g., 10-K, 10-Q, 8-K), 44 formal financial statements, 25 securities transaction records, and 11 tax forms (e.g., Form 4, Form 990). Crucially, we prioritized pages exhibiting extreme visual complexity: heterogeneous tabular structures, dense numerical blocks, cross-page charts, and long-form narratives embedded with microscopic footnotes. 

\subsubsection{Financial Fact Annotation}

To construct reliable factual supervision, four trained annotators with finance, accounting, and computer science backgrounds manually labeled the ground-truth HTML with the five types of financially critical facts defined in Table~\ref{tab:fact_types}, contributing over 150 expert annotation hours. Following a rigorous, pilot-refined annotation guideline, annotators precisely marked the exact minimal text spans corresponding to each fact using the Label Studio platform. By strictly excluding auxiliary tokens, this precise span-level annotation ensures deterministic alignment between OCR predictions and the underlying financial facts. Detailed annotation guideline and the Label Studio interface are provided in the Appendix.


\subsubsection{Annotation Quality and Statistics}

We validated annotation rigor using both pairwise (Cohen’s~$\kappa$) and multi-rater (Fleiss’~$\kappa$) agreement metrics. Pairwise scores ranged from 0.83 to 0.93, culminating in a robust overall Fleiss’~$\kappa$ of 0.8837 (Table~\ref{tab:agreement}), confirming high inter-annotator consistency and overall dataset quality. The finalized \ficriticaled dataset comprises \textbf{859} samples yielding 9,481 annotated facts, averaging 11 per document (Table~\ref{tab:dataset_stats}). Notably, over 48\% of the samples feature an intersection of multiple critical fact types, reflecting the dense, multimodal entanglement characteristic of real-world financial reporting.

\begin{table}[t]
\setlength{\abovecaptionskip}{3pt}
\setlength{\belowcaptionskip}{0pt}
\caption{Inter-annotator agreement measured by pairwise Cohen’s~$\kappa$ and overall Fleiss’~$\kappa$. 
}
\label{tab:agreement}
\centering
\scriptsize
\renewcommand{\arraystretch}{0.85}

\begin{tabular}{lcccc}
\toprule
\textbf{Annotator} & \textbf{1} & \textbf{2} & \textbf{3} & \textbf{4} \\
\midrule
1 & ---    & 0.8789 & 0.8952 & 0.9159 \\
2 & 0.8789 & ---    & 0.8511 & 0.8294 \\
3 & 0.8952 & 0.8511 & ---    & 0.9334 \\
4 & 0.9159 & 0.8294 & 0.9334 & ---    \\
\midrule
\textbf{Overall (Fleiss’~$\kappa$)} & \multicolumn{4}{c}{\textbf{0.8837}} \\
\bottomrule
\end{tabular}

\end{table}
\begin{table}[t]
\setlength{\abovecaptionskip}{3pt}
\setlength{\belowcaptionskip}{0pt}
\caption{Statistics and financial critical facts distribution of the \ficriticaled dataset.}
\centering
\label{tab:dataset_stats}
\resizebox{\linewidth}{!}{
\begin{tabular}{lccc}
\toprule
\textbf{Type} & \textbf{Count} & \textbf{Average per Document} & \textbf{Percentage (\%)} \\
\midrule
Dataset Size & \textbf{859} & -- & -- \\
\textit{Numeric Facts} ($F_n$) & 2,930 & 3.4 & 30.9 \\
\textit{Temporal Facts} ($F_t$) & 1,144 & 1.3 & 12.1 \\
\textit{Monetary Units} ($F_{mu}$) & 881 & 1.0 & 9.3 \\
\textit{Reporting Entities} ($F_{re}$) & 3,142 & 3.7 & 33.1 \\
\textit{Financial Concepts} ($F_{fc}$) & 1,384 & 1.6 & 14.6 \\
Total ($F$) & \textbf{9,481} & \textbf{11.0} & \textbf{100.0} \\
\bottomrule
\end{tabular}}
\end{table}

\subsection{Financial Fact Evaluation}

Traditional OCR evaluation relies on surface-level lexical metrics (e.g., ROUGE, edit distance) that reward superficial string overlap. However, these metrics fail to assess whether a model preserves the \textit{financial meaning} of a document. In finance, even single-character OCR deviations can alter the underlying reality. \ficriticaled operationalizes a shift from text similarity to \textit{fact-level correctness}, measuring the exact preservation of domain-critical factual content.


\subsubsection{Financial Fact Accuracy (FFA) Formulation}

The objective is to verify whether a model correctly reproduces the annotated facts $F = F_n \cup F_t \cup F_{mu} \cup F_{re} \cup F_{fc}$ from the ground-truth HTML within the model-generated output. For each fact $f_i \in F$, we assign a binary correctness indicator $\delta_i \in \{0, 1\}$, where $\delta_i = 1$ if $f_i$ is recovered in a semantically equivalent local context, and $0$ otherwise. Crucially, factual matching across all categories requires the absolute preservation of visually subtle but determinant cues, including signs, decimals, unit scales, temporal boundaries, and entity attribution. For example, converting ``(1,200)'' to ``1,200'' yields $\delta_i = 0$, as the omission of the parenthetical negative notation fundamentally inverts the financial meaning.

For a given output document, the overall Financial Fact Accuracy (\textbf{FFA}) is computed as $\alpha = \sum_i \delta_i / |F|$. We similarly define category-specific accuracies as $\alpha_x = \sum_{f_i \in F_x} \delta_i / |F_x|$ for each subset $x \in \{n, t, mu, re, fc\}$. To obtain robust dataset-level performance normalized across documents with varying fact densities, we aggregate the total correct counts $C_x$ and total fact counts $T_x$ across all samples. The global benchmark metric for each category is thus defined as $\text{x-FFA} = C_x / T_x$, yielding distinct evaluations for numeric (\textbf{N-FFA}), temporal (\textbf{T-FFA}), monetary unit (\textbf{MU-FFA}), reporting entity (\textbf{RE-FFA}), and financial concept (\textbf{FC-FFA}) preservation. The overall dataset-level \textbf{FFA} is computed analogously over all facts in $F$, providing a unified measure of factual fidelity.

\subsubsection{Deterministic-Rule-Guided LLM-as-Judge Verification}

Because financial facts do not map cleanly onto surface forms, traditional string overlap is fundamentally insufficient for evaluation. A single fact may admit multiple valid expressions (e.g., ``\$1.2B'' vs.\ ``1,200 million''), whereas a lexically minor deviation can entirely falsify the preserved information (e.g., ``16,950'' vs.\ ``16.950''). To evaluate at the fact level, we employ a \textit{Deterministic-Rule-Guided LLM-as-Judge} pipeline combining deterministic parsing with LLM-powered semantic judgment across the five fact types. The pipeline proceeds in three stages: (1) Extract all annotated entities from
ground-truth HTML with $\pm$40-character context hints for disambiguation; (2) LLM--as-Judge framework (powered by GPT-4o) determines per entity whether it was \textit{found} and \textit{exactly matched} in the prediction; and (3) any
entity passing a normalized exact-match check is deterministically forced to $\text{correct} = \texttt{true}$, guarding against LLM false negatives. Rather than penalizing superficial formatting differences, the judge focuses on semantic fidelity: it determines whether the model has successfully located and reproduced each entity, and whether the reproduced value is factually exact. Conditioned on a structured evaluation prompt (see in Appendix), the Deterministic-Rule-Guided LLM-as-Judge outputs per-field correctness decisions used to compute accuracy metrics across critical fields types.

To rigorously validate the automated evaluator, we conducted a \textit{human--LLM alignment validation} by sampling 15 model outputs each from PP-OCRv5, GPT-4o, and GPT-5. We adopt a stringent \textit{Document-Level Strict Agreement} metric: the LLM judge and human expert are considered in agreement if and only if their correctness decisions ($\delta_i$) match perfectly across \textit{all} entities within a document. Even a single entity-level divergence results in a document-level disagreement. Following iterative prompt and workflow refinement, the evaluation protocol achieves a robust 95\% agreement rate, confirming its reliability as a domain-aware judge for complex financial factual verification.







\subsection{Experimental Setup}

\subsubsection{Evaluation Models}
We evaluate four categories of systems to comprehensively assess financially critical document understanding:

\noindent \textbf{OCR Pipelines.}
Modular systems that decouple text detection, recognition, and document reconstruction, including \textit{MinerU2.5}~\cite{niu2025mineru25decoupledvisionlanguagemodel} and \textit{PP-OCRv5}~\cite{cui2025paddleocr30technicalreport}. These serve as strong engineering-oriented baselines for structured document parsing.

\noindent \textbf{Specialized OCR VLMs.}
Specialized foundation models optimized for visually grounded text extraction and structure-aware recognition, including \textit{DeepSeek-OCR}~\cite{wei2025deepseek}, \textit{DeepSeek-OCR-2}~\cite{wei2026deepseek}, and GLM-OCR~\cite{duan2026glmocrtechnicalreport}. These provide a tightly integrated OCR-centric alternative to traditional pipelines.

\noindent \textbf{Open-Source MLLMs.}
Unified multimodal models that jointly perform visual understanding and language reasoning, including \textit{Gemma-3n-E4B-it}~\cite{google_gemma_3n_E4B_it_2025}, \textit{Qwen3-VL-8B-Instruct}~\cite{qwen3technicalreport}, \textit{Llama-4-Mav-} \textit{erick}~\cite{meta-llama-4-maverick-17b-128e-instruct-fp8}, and \textit{Qwen3.5-397B-A17B}~\cite{qwen3.5}.

\noindent \textbf{Proprietary MLLMs.}
Frontier closed-source models with strong multimodal reasoning capabilities, including \textit{GPT-4o}~\cite{hurst2024gpt}, \textit{GPT-5}~\cite{openai_gpt5}, \textit{Claude-Sonnet-4.6}~\cite{anthropic2026claude46}, and \textit{Gemini-2.5-Pro}~\cite{gemini25pro}. These serve as upper-bound references for end-to-end document understanding without external OCR.

\subsubsection{Implementation Details}
All proprietary models are accessed via their official APIs with temperature set to zero for deterministic outputs. Open-source MLLMs are hosted by TogetherAI API. OCR pipelines and specialized OCR VLMs are either self-hosted on four NVIDIA GPUs or accessed via their public API endpoints. All models share a unified input--output format aligned with the \ficriticaled annotation schema. 

\section{Results and Analysis}

To assess whether current OCR and MLLM systems truly preserve financially critical, visually grounded evidence, we organize our experiments around the following three research questions. 

\begin{table}[ht]
\setlength{\abovecaptionskip}{3pt}
\setlength{\belowcaptionskip}{0pt}
\caption{
Model performances for traditional OCR pipelines, specialized OCR VLMs, open-source MLLMs, and proprietary MLLMs. 
\textbf{Bold} indicates the best performance in each column, while \underline{underlined} values denote the second-best. For error rate (E$\downarrow$), lower is better. 
Fact-level metrics include numerical (N-FFA), temporal (T-FFA), monetary unit (MU-FFA), reporting entity (RE-FFA), and financial concept (FC-FFA) accuracy; FFA denotes overall financial fact accuracy.
}
\label{tab:exp_res}
\centering
\setlength{\tabcolsep}{4pt}
\renewcommand{\arraystretch}{0.97}
\resizebox{\linewidth}{!}{%
\begin{tabular}{l|c|ccc|cccccc}
\toprule
\textbf{Model} 
& \textbf{Size}
& \multicolumn{3}{c|}{\textbf{General (\%)}}
& \multicolumn{6}{c}{\textbf{Fact-Level (\%)}} \\
\cmidrule(lr){3-5} \cmidrule(lr){6-11}
& 
& ROUGE-1$\uparrow$ & ROUGE-L$\uparrow$ &  Edit Distance$\downarrow$
& N-FFA$\uparrow$ & T-FFA$\uparrow$ & MU-FFA$\uparrow$ & RE-FFA$\uparrow$ & FC-FFA$\uparrow$ & FFA$\uparrow$ \\
\midrule
\rowcolor{finrow} \multicolumn{11}{c}{\textit{OCR Pipelines}} \\
MinerU2.5 ~\cite{niu2025mineru25decoupledvisionlanguagemodel}
& 1.2B 
& 95.71 & 95.30 & 6.02
& \textbf{98.76} & 96.48 & 54.05 & 91.09 & 96.44 & 94.64 \\
PP-OCRv5 ~\cite{cui2025paddleocr30technicalreport}
& 0.07B
& 97.54 & 96.55 & 3.10 
& 95.70 & 90.29 & 90.00 & 86.62 & 93.75 & 91.91 \\
\rowcolor{finrow} \multicolumn{11}{c}{\textit{Specialized OCR VLMs}} \\
DeepSeek-OCR ~\cite{wei2025deepseek}
& 6B
& 94.73 & 94.42 & 7.33 
& 93.47 & 91.96 & 83.53 & 92.27 & 94.36 & 92.67 \\
DeepSeek-OCR-2 ~\cite{wei2026deepseek}
& 3B
& 92.90 & 92.18 & 10.72 
& 82.63 & 91.90 & 82.83 & 88.69 & 86.51 & 86.19 \\
GLM-OCR ~\cite{duan2026glmocrtechnicalreport}
& 0.9B
& 95.10 & 94.74 & 6.43 
& 93.24 & \textbf{98.53} & 88.89 & \textbf{97.84} & \textbf{100.00} & \underline{96.92} \\
\midrule
\rowcolor{finrow} \multicolumn{11}{c}{\textit{Open-source MLLMs}} \\
Gemma-3n-E4B-it ~\cite{google_gemma_3n_E4B_it_2025}
& 4B
& 83.49 & 79.59 & 23.82 
& 52.65 & 77.06 & 64.71 & 74.65 & 72.86 & 65.68 \\
Qwen3-VL-8B-Instruct ~\cite{qwen3technicalreport}
& 8B
& 97.68 & 97.40 & 2.93 
& \underline{98.47} & 96.99 & \textbf{97.65} & 93.18 & \underline{99.24} & 96.88 \\
Llama-4-Maverick ~\cite{meta-llama-4-maverick-17b-128e-instruct-fp8}
& 17B
& 98.00 & 97.62 & 3.70 
& 97.77 & \underline{97.99} & \textbf{97.65} & \underline{94.26} & 98.48 & \underline{96.92} \\
Qwen3.5-397B-A17B ~\cite{qwen3.5}
& 397B
& \underline{98.12} & \underline{98.00} & 2.59 
& 87.72 & 87.99 & 86.14 & 91.22 & 94.40 & 89.70 \\
\midrule
\rowcolor{finrow} \multicolumn{11}{c}{\textit{Proprietary MLLMs}} \\
GPT-4o ~\cite{hurst2024gpt}
& -
& 90.40 & 88.35 & 16.01 
& 59.56 & 84.59 & 81.92 & 81.78 & 70.84 & 71.68 \\
GPT-5 ~\cite{openai_gpt5}
& -
& 91.81 & 89.56 & 15.79 
& 66.83 & 94.48 & 92.35 & 89.19 & 91.77 & 81.65 \\
Claude-Sonnet-4.6 ~\cite{anthropic2026claude46}
& -
& \textbf{98.84} & \textbf{98.73} & \textbf{1.69} 
& \underline{98.59} & \underline{97.99} & \underline{97.06} & 94.02 & 98.94 & \textbf{97.23} \\
Gemini-2.5-Pro ~\cite{gemini25pro}
& -
& \underline{98.81} & \underline{98.37} & \underline{2.46} 
& 97.24 & 97.82 & \textbf{97.65} & 94.18 & 98.94 & 96.74 \\
\bottomrule
\end{tabular}%
}
\end{table}

\subsection{Main Results and Discussion}

\subsubsection{General OCR Metrics vs. Financial Fidelity}
\textbf{RQ1:} To what extent do conventional OCR metrics reflect financially faithful extraction of critical fields?

The results show that conventional OCR metrics are insufficient for evaluating financial fidelity. While they capture overall reconstruction quality, they do not reliably reflect whether financially critical fields are preserved faithfully. 
For example, MinerU2.5 achieves a near-ceiling ROUGE-1 score of 95.71\% and ROUGE-L score of 95.30\%, yet drops sharply to 54.05\% on MU-FFA, indicating that strong lexical overlap can still mask serious fact-level errors. Similar gaps remain even among the strongest models: Claude-Sonnet-4.6 reaches 98.84\% ROUGE-1 and 98.73\% ROUGE-L but only 94.02\% on RE-FFA, while Qwen3.5-397B-A17B attains 98.12\% ROUGE-1 and 98.00\% ROUGE-L yet falls to 86.14\% on MU-FFA. These results show that high overlap-based OCR scores do not necessarily imply financially faithful extraction, especially when correctness depends on exact values, units, reporting subjects, and field associations.

\subsubsection{Specialized OCR Models vs. General MLLMs}
\textbf{RQ2:} How do specialized OCR models and general-purpose MLLMs differ in their ability to preserve financially critical information?

In our benchmark, OCR-specialized models show strong performance but not uniformly dominant on financially critical extraction. Among the OCR-specialized systems, GLM-OCR achieves 96.92\% FFA, outperforming PP-OCRv5 at 91.91\% and DeepSeek-OCR-2 at 86.19\%, which indicates that OCR-oriented models remain competitive baselines for exact field preservation. 
At the same time, several general MLLMs achieve comparable or stronger fact-level performance, including Qwen3-VL-8B-Instruct at 96.88\%, Llama-4-Maverick at 96.92\%, Gemini-2.5-Pro at 96.74\%. Among these models, Claude-Sonnet-4.6 gets the best result with 97.23\%. 
It shows the great potential of MLLMs in OCR tasks. However, this capability is not consistent across the broader general-MLLM group: Gemma-3n-E4B-it reaches only 65.68\% FFA, GPT-4o 71.68\%, and GPT-5 81.65\%, all below the strongest OCR-oriented and multimodal systems. Overall, these findings suggest that OCR specialization remains an important advantage for financially faithful transcription, although the strongest general MLLMs can now match or even exceed it.

\begin{figure}[t]
\setlength{\abovecaptionskip}{2pt}
\setlength{\belowcaptionskip}{0pt}
\centering
\includegraphics[width=0.96\linewidth]{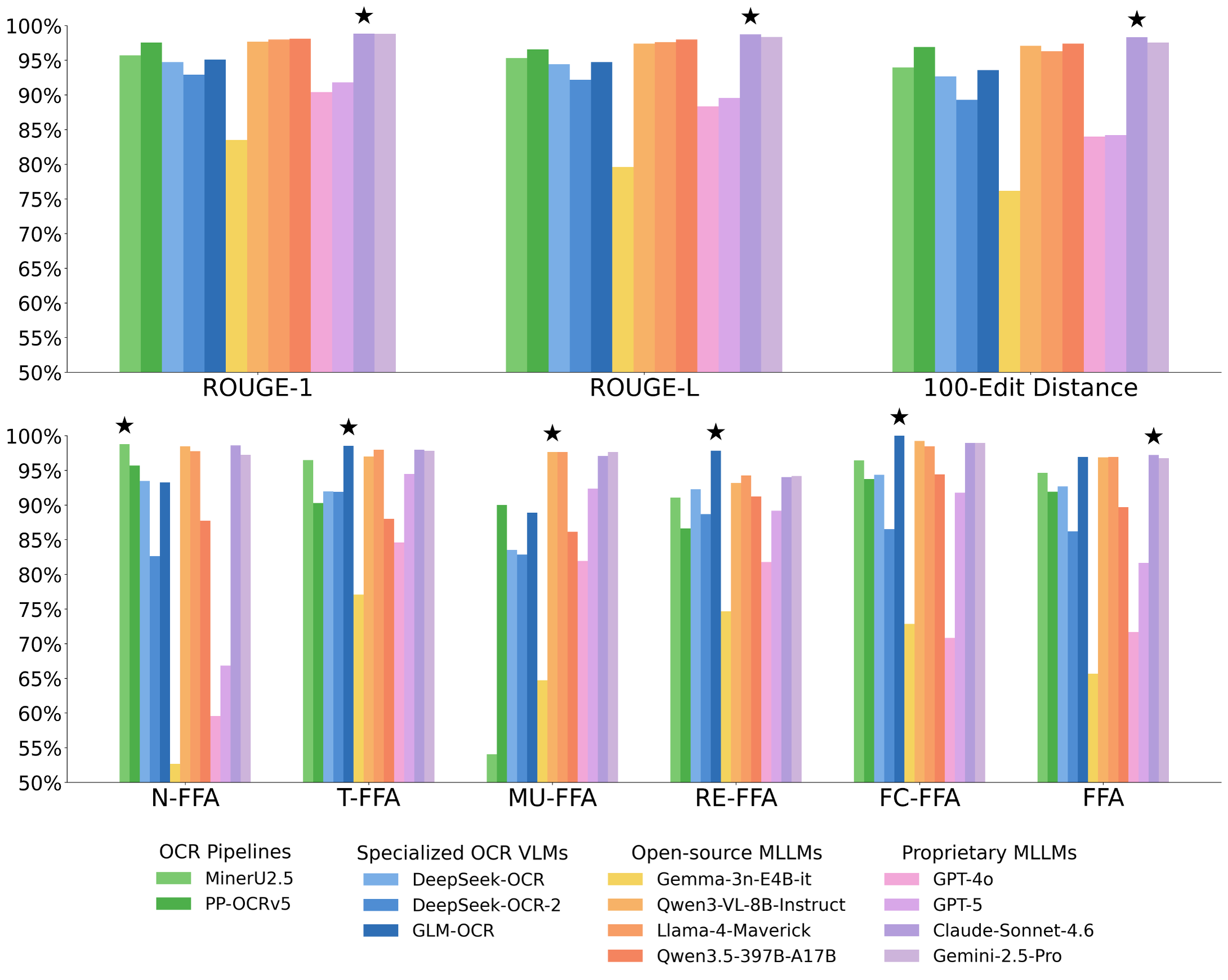}
\caption{
Model performance on general OCR metrics and financial fact-level accuracies.
Best results per metric are highlighted with a star (\(\star\)).
}
\label{fig:result}
\end{figure}

\subsubsection{Difficulty Across Financial Critical Fact Types}
\textbf{RQ3:} Which types of financial critical facts are most vulnerable to OCR errors across current models?

The results show that numerical values and monetary units are the most vulnerable fact types across current models, while temporal facts and financial concepts perform relatively better but remain far from robust. For example, Gemma-3n-E4B-it achieves only 52.65\%, GPT-4o 59.56\%, and GPT-5 66.83\% on numerical facts, indicating a substantial gap in preserving reliable numeric accuracy—an essential requirement for financial document understanding. MinerU2.5 reaches only 54.05\% on MU-FFA, suggesting that nearly half of monetary unit fields are incorrectly recognized, which can severely distort financial meaning. Even Claude-Sonnet-4.6, despite its more balanced overall performance, still falls short of the near-perfect accuracy required for real-world financial applications.
Notably, even the comparatively stronger fact categories fail to reach trustworthy levels. For instance, Qwen3.5-397B-A17B achieves 87.99\% on T-FFA and 94.40\% on FC-FFA, while weaker models such as Gemma-3n-E4B-it drop to 52.65\% on N-FFA and 77.06\% on T-FFA.
Overall, these results suggest that compact local cues, such as digits, decimal points, signs, and unit markers, are the most error-sensitive elements in financial OCR, and that even relatively stronger fact types have not yet achieved the level of reliability required for high-stakes financial interpretation.

\subsection{Model Failure Analysis}
\label{sec:error-analysis}

Beyond aggregate fact-level accuracy, we conduct a page-level error analysis of \textbf{306} OCR cases from \textbf{6} representative models, all reviewed by financial experts. 
While FFA evaluates whether annotated \emph{financial critical facts} are preserved, expert review further captures \emph{non-critical OCR errors, layout distortions, and other page-level failures} that still affect practical use. 
This means that a model may achieve high FFA or ROUGE scores yet still exhibit substantial page-level errors.
We analyze these failures along three axes: (\textit{i}) \textbf{financial severity}, distinguishing between critical-field and non-critical OCR errors; (\textit{ii}) \textbf{document modality and complexity}, examining how performance varies across page types and difficulty levels; and (\textit{iii}) \textbf{model-specific failure patterns}, characterizing recurring error behaviors across model families.

\paragraph{\textbf{Financial severity}}
We define a \textbf{critical error} as any OCR mismatch involving one of the five annotated critical field types, and a \textbf{non-critical error} as any OCR discrepancy outside these fields, such as errors in surrounding narrative text, punctuation, or spacing. Table~\ref{tab:page_level_error_stats} shows large differences in page-level financial reliability: the strongest systems make critical errors on only 17.6\%--23.5\% of pages, while weaker baselines do so on nearly three-fourths of the benchmark. Non-critical OCR errors remain common across almost all models, ranging from 49.0\% to 90.2\%, showing that strong fact preservation does not imply clean transcription.

\begin{table}[t]
\setlength{\abovecaptionskip}{3pt}
\setlength{\belowcaptionskip}{0pt}
\centering
\caption{Page-level error statistics across models. Critical error: if at least one annotated financially critical field is incorrect; non-critical error: if OCR discrepancies occur outside annotated critical spans.}
\label{tab:page_level_error_stats}
\resizebox{\linewidth}{!}{
\begin{tabular}{l|cc|cccc}
\toprule
\textbf{Model} & \textbf{Critical Error} & \textbf{Non-Critical Error} & \textbf{Clean Pages} & \textbf{Critical Only} & \textbf{Non-Critical} & \textbf{Both} \\
 & \textbf{Rate (\%) $\downarrow$} & \textbf{Rate (\%) $\downarrow$} & \textbf{(\%) $\uparrow$} & \textbf{(\%) $\downarrow$} & \textbf{Only (\%) $\downarrow$} & \textbf{(\%) $\downarrow$} \\
\midrule
\rowcolor{finrow} \multicolumn{7}{c}{\textit{OCR Pipeline}} \\
MinerU2.5~\cite{niu2025mineru25decoupledvisionlanguagemodel}      & 23.5 & 58.8 & 32.4 & 8.8 & 44.1 & 14.7 \\
\midrule
\rowcolor{finrow} \multicolumn{7}{c}{\textit{Specialized OCR VLMs}} \\
DeepSeek-OCR ~\cite{wei2025deepseek}   & 17.6 & 86.3 & 13.7 & 0.0 & 68.6 & 17.7 \\
\midrule
\rowcolor{finrow} \multicolumn{7}{c}{\textit{Open-source MLLMs}} \\
Gemma-3n-E4B-it~\cite{google_gemma_3n_E4B_it_2025}      & 74.5 & 90.2 & 7.8 & 2.0 & 17.7 & 72.6 \\
Llama-4-Maverick ~\cite{meta-llama-4-maverick-17b-128e-instruct-fp8}
       & 15.7 & 49.0 & 49.0 & 2.0 & 35.3 & 13.7 \\
\midrule
\rowcolor{finrow} \multicolumn{7}{c}{\textit{Proprietary MLLMs}} \\
GPT-4o ~\cite{hurst2024gpt}& 52.9 & 80.4 & 11.8 & 7.8 & 35.3 & 45.1 \\
GPT-5 ~\cite{openai_gpt5} & 58.0 & 86.0 & 12.0& 2.0 & 30.0& 56.0 \\

\bottomrule
\end{tabular}}
\end{table}

A decomposition into \textit{clean}, \textit{non-critical only}, \textit{critical only}, and \textit{both} further highlights this gap. Llama-4-Maverick yields the highest share of clean pages (49.0\%), whereas Gemma-3n-E4B-it has the largest share of pages containing both critical and non-critical errors (72.6\%). Thus, practical financial OCR requires minimizing both materially dangerous errors and broader transcription artifacts.

\paragraph{\textbf{Document modality and complexity}}
We next stratify page-level critical error rates by document modality and visual complexity. Pages are grouped into \textit{text-only}, \textit{table-only}, and \textit{mixed} modalities, and into \textit{Low}, \textit{Medium}, and \textit{High} complexity levels (definitions are in the Appendix).

As shown in Figure~\ref{fig:complexity_modality_heatmaps}, \textit{mixed} pages are consistently the hardest setting. Weaker general-purpose MLLMs exhibit very high critical error rates in this category, with Gemma-3n-E4B-it reaching 100.0\% across all complexity levels and GPT models remaining at 80.0\%--100.0\%. By contrast, OCR-specialized models are more stable, though not error-free: on \textit{text-only} pages, DeepSeek-OCR rises from 0.0\% at low complexity to 23.1\% at high complexity, and MinerU2.5 rises from 0.0\% to 38.5\%. Table-only pages are generally easier when layouts are simple. Overall, the main challenge is OCR under heterogeneous structure rather than OCR alone.

\begin{figure}[t]
\setlength{\abovecaptionskip}{8pt}
\setlength{\belowcaptionskip}{0pt}
\centering
\includegraphics[width=1.0\linewidth]{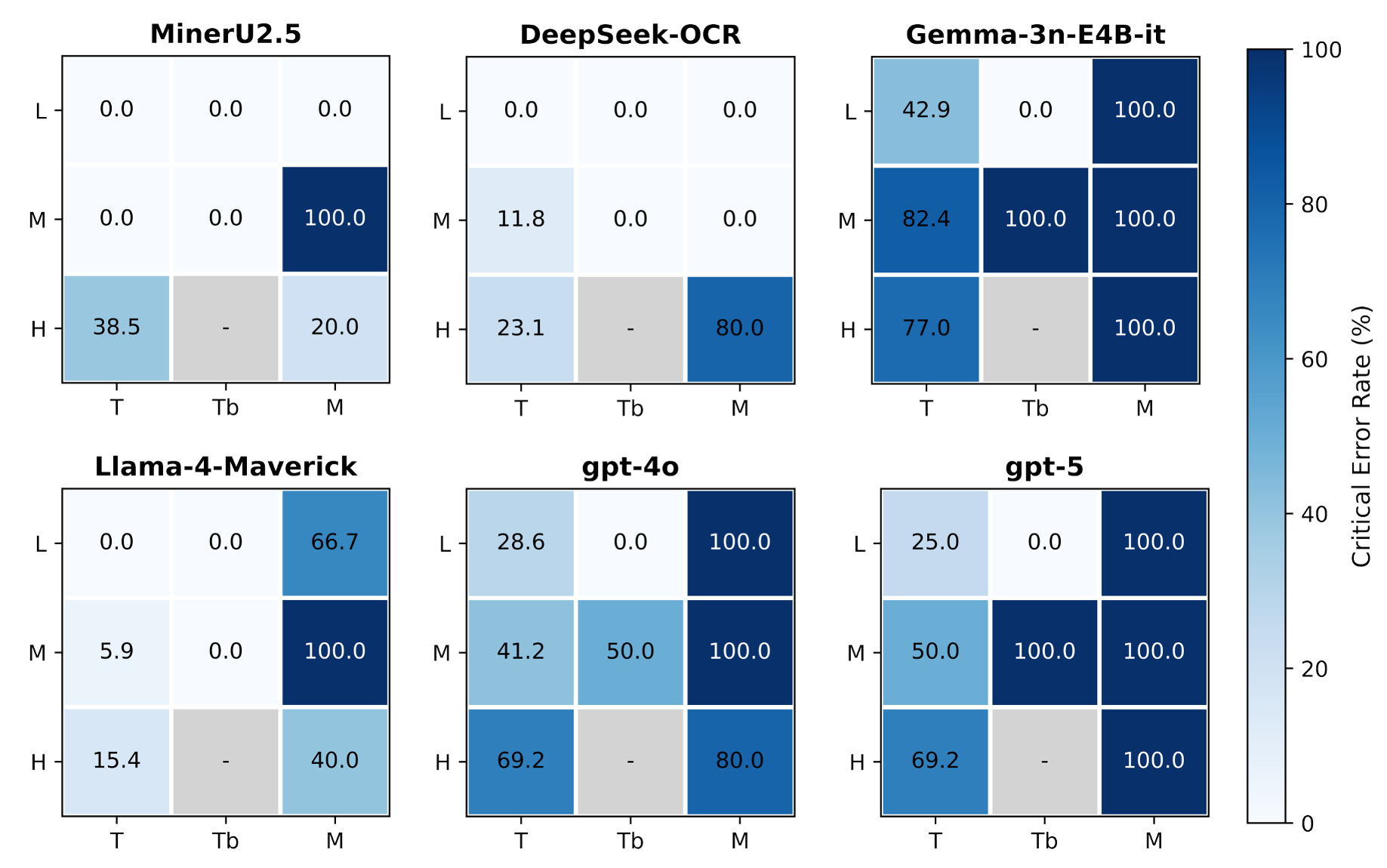}
\caption{
Page-level critical error rate (\%) by document complexity and modality. Each subtable reports a 3$\times$3 breakdown where rows denote complexity (Low, L; Medium, M; High, H) and columns denote modality (text-only, T; table-only, Tb; mixture, M). Grey cell with "-" represents 0 sample under this category.
}
\label{fig:complexity_modality_heatmaps}
\end{figure}

\paragraph{\textbf{Model-specific failure patterns}}
Beyond aggregate error rates, model families exhibit distinct failure modes. OCR-specialized systems such as DeepSeek-OCR and MinerU2.5 mostly make fixed, repetitive recognition errors, including recurring lexical corruptions (e.g., \emph{transferor} $\rightarrow$ \emph{transfessor}) and stable formatting artifacts around punctuation or dashes. This is consistent with their architectures: MinerU2.5 uses a coarse-to-fine two-stage pipeline that separates layout analysis from local recognition~\citep{niu2025mineru25decoupledvisionlanguagemodel}, while DeepSeek-OCR relies on optical compression to represent high-resolution pages with compact visual tokens~\citep{wei2025deepseek}. By contrast, weaker or more generative multimodal LLMs produce more diverse and less predictable errors. In Gemma-3n-E4B-it and GPT-4o, we observe \textbf{hallucinated continuation}, where the model incorrectly completes a sentence using previously seen text with a similar prefix, along with context-driven substitutions and unstable free-form corruption~\citep{google_gemma_3n_E4B_it_2025,hurst2024gpt}. Strong general models such as Llama-4-Maverick remain more stable at the page level, suggesting a distinction between grounded transcription and semantic reconstruction.

\section{Conclusion}

In this work, we introduced \ficriticaled, a fact-centric visual benchmark for evaluating whether OCR systems and MLLMs preserve financially critical evidence in real-world documents. 
The benchmark includes an expert-annotated dataset of financially critical fields and an LLM-powered evaluation protocol for structured fact-level assessment.
Our experiments yield three main findings: (1) Conventional text similarity metrics do not reliably capture financial faithfulness: high surface-level overlap can still mask substantial fact-level errors. (2) Not all critical fields are equally difficult: numerical fields and monetary units are consistently more fragile than temporal, reporting, and concept fields, showing that the main challenge lies in preserving small local cues such as digits, signs, and unit markers. (3) OCR-specialized systems remain strong and stable baselines, but the strongest MLLMs can now match or exceed them, suggesting that grounded transcription behavior matters more than model family alone.
Overall, \ficriticaled reframes financial OCR as a fact-preservation problem rather than a surface transcription task. We hope the benchmark will support future research on financially faithful, layout-aware document understanding in precision-critical domains.

\clearpage
\bibliographystyle{ACM-Reference-Format}
\bibliography{fincriticaled_bib}

\clearpage
\appendix
\section{Dataset Construction Details}
We provide the complete pipeline for data collection, preprocessing, and annotation. This includes document source distribution, sampling criteria, and anonymization protocols. Each document is stripped of personally identifiable information and re-rendered into page-level PDFs for OCR benchmarking.

\paragraph{Document Sources and Sampling}
Documents are collected from publicly available financial repositories, including the U.S. Securities and Exchange Commission (SEC) and open-source tax forms for non-profit organizations. We include five primary document types: \textit{financial statements}, \textit{required SEC filings}, \textit{tax forms}, \textit{securities transaction records}, and \textit{financial legal documents}. To ensure annotation quality, each sampled filing must contain at least one type of critical fields. 

\begin{table}[ht]
\setlength{\abovecaptionskip}{3pt}
\setlength{\belowcaptionskip}{0pt}
\caption{
Financial Documents Type Distribution
}
\label{tab:coverage}
\centering
\setlength{\tabcolsep}{4pt}
\renewcommand{\arraystretch}{0.97}
\resizebox{\linewidth}{!}{%
\begin{tabular}{p{2cm}cp{3cm}p{4cm}c}
\toprule
\textbf{Document Type} 
& \textbf{Data Source}
&\textbf{Examples}
& \textbf{Characteristics} 
& \textbf{Data size}
\\
\midrule
Financial Statements
& SEC EDGAR 
& 10-K, 10-Q 
& Dense tables, mixture of 
modality and formats & 44 \\
\midrule
SEC Required filings
& SEC EDGAR 
& Form 4
, Form 8-K
& Complex table structure & 142 \\
\midrule
Tax forms
& Candid
& Form 990
& Cross page tables and charts & 11 \\
\midrule
Transaction Records
& SEC EDGAR 
& Securities Transaction 
Records
& Well structured table, 
numerically heavy & 25 \\
\midrule
Financial Legal Documents
& SEC EDGAR 
& M\&A agreements, 
credit agreements, 
bond indentures
& Long-form legal text with embedded financial 
tables, contract terms & 637 \\
\bottomrule
\end{tabular}%
}
\end{table}

\paragraph{Preprocessing and Rendering}
All source documents are converted into canonical HTML using a rule-based extraction pipeline. Each HTML file is then paired with its corresponding page-level image deriving from intermediate PDFs, rendered at 300~dpi using Chromium headless mode to preserve layout fidelity. During rendering, embedded objects such as charts, tables, and formulas are retained as rasterized elements. For each document, metadata (document type, filing year) is recorded and stored in structured JSON format. To ensure compatibility with OCR models, we further normalize text alignment, table borders, and visual regions.

\paragraph{Annotation and Quality Control}
Annotations are performed directly on HTML text, using the corresponding page image as visual reference for validation. Annotators highlight entities within HTML and embed specialized span tags encoding the five critical value types. An internal agreement table reports consistent reliability across both entity types (see Table~\ref{tab:agreement} and appendix \ref{sec:data_annotation}).


\paragraph{Final Statistics and Accessibility}
The final dataset contains 859 fully annotated financial documents, totaling 9,481 annotated entities.Each page is accompanied by its source metadata, HTML text, and annotated span-level markup. All files are distributed under a research-only license via a secure hosting repository.

\section{Data Annotation}
\label{sec:data_annotation}

\ficriticaled annotation procedure revolves around entity labeling for financial documents. The goal is to identify financially critical entities, focusing on numbers and time.


\reviewhead{Annotation Guideline}
\begin{Verbatim}[
  fontsize=\scriptsize,
  frame=single,
  rulecolor=\color{black},
  framesep=3mm,
  labelposition=topline
]
Entity List
- Number
- Temporal
- Monetary Unit
- Reporting Entity
- Financial Concept

General Rules
- Identify all entities in the HTML file that belong to one of the five
  categories listed above.
- Use the rendered HTML as the primary source of truth; use the corresponding
  page image only for visual assistance when the layout or OCR text appears
  ambiguous.
- Highlight all valid entities in each task without omission.

Guidelines on entity type definition and label instructions
- Number
  - The numeric value should be financially critical, including percentages,
    monetary amounts, and share amounts.
  - Annotate only the number itself, or the number with attached signs, which
    may include decimal points, commas for thousands, and negative markers.
  - Do not include non-financially-related numbers, such as serial numbers or
    policy section numbers.
  - Examples: 1,000,000, 2,345, 0.37, 10, 1/3, -2.3, (10,234), 25.63%.
    Do not annotate examples such as Section 30(h).
- Temporal (Duration and Dates)
  - Only include specific dates and time periods. Do not include time 
  frequency expressions such as monthly or weekly.
  - If the annotation target is a date range, annotate only the date itself 
  and remove any context words between the two dates.
  - Examples: March 24, 2025, 1 month, and 2 years should be annotated.
  - For a span such as PERIODS JANUARY 2, 2025 THRU JANUARY 8, 2025, only
    annotate JANUARY 2, 2025 and JANUARY 8, 2025; exclude PERIODS and THRU.
- Monetary Unit
  - Monetary-related units should be included.
  - Examples: $, USD, thousands, millions, share.
- Reporting Entity
  - If the entity, such as a company name, contains the full name followed 
  by the abbreviation, annotate them together.
  - Example: 1st Franklin Financial Corporation ("1FFC").
  - Additional examples: 1st FRANKLIN FINANCIAL CORPORATION, 1832 Asset
    Management L.P.
  - Include the whole entity, e.g., American Century Investment Management,
    Inc.
- Financial Concepts
  - Only label financial concepts if they appear inside a table or form. 
  There is no need to label table titles or footnotes.
  - Label the whole line item if it contains a financial concept.
  - Example: Maximum Sales Charge (Load) Imposed on Purchases (as a 
  percentage of offering price).
  - Label a term as Financial Concept if it belongs to one of the 
  following categories:
  
    1. Revenue and Sales
       Concepts describing inflows from core business activities.
       Examples: Revenue, Net revenue, Gross revenue, Sales, Net sales,
       Service revenue, Total Revenue.
    2. Income, Profit, and Earnings
       Concepts describing profitability or earnings outcomes.
       Examples: Income, Net income, Operating income, Gross profit, 
       Profit, Earnings, Earnings per share (EPS).
    3. Costs, Expenses, and Losses
       Concepts describing outflows, reductions, or negative performance.
       Examples: Cost, Cost of revenue, Cost of sales, Operating expenses,
       Expense, Loss, XXX Fee, XXX Expenses.
    4. Taxes
       Concepts related to taxation and tax-related outcomes.
       Examples: Income tax, Tax expense, Deferred tax, Tax liability,
       Effective tax rate.
    5. Margins and Ratios
       Common, high-level ratios directly tied to revenue or profit.
       Examples: Gross margin, Operating margin, Profit margin,
       Net Income/Loss.
    6. Financial Obligations and Commitments
       Concepts describing required or expected payments.
       Examples: Lease obligation, Purchase obligation, Debt,
       Interest expense, Interest income.
  - Explicit Exclusions (Do NOT label)
    1. Operational or Functional Activities:
       Research and Development, Marketing, Sales and distribution,
       General and administrative.
    2. Accounting or Reporting Metadata:
       Accounting policy, Notes to the financial statements,
       Segment information, Management discussion and analysis (MD&A).

    3. Broad Business or Strategy Terms:
       Growth strategy, Market expansion, Customer acquisition,
       Product roadmap.
    4. Non-Financial Metrics or Qualitative Descriptions:
       Headcount, Employee engagement, ESG initiatives.
\end{Verbatim}

\subsection{Annotator Demography}
\label{sec:sec_annotator_demography}

\ficriticaled was annotated by a four-person team with complementary expertise in finance, economics, auditing, and computer technology. The team combines professional experience in financial analysis and FinTech with graduate-level training in business analytics, financial mathematics, computer science, and computer technology.

One annotator is a principal analyst at a major U.S. financial institution with training in business analytics, statistics, and economics, as well as experience in LLMs, financial data analysis, and multilingual reasoning. Another annotator has training in financial mathematics and computer science, along with more than seven years of experience in strategic finance and consulting in the FinTech industry. The remaining two annotators are graduate students in computer technology with backgrounds in auditing, financial analysis, data processing, annotation workflows, and LLM evaluation and adaptation for domain-specific tasks.

Overall, the team reflects a balance between financial domain knowledge and technical expertise, supporting accurate, consistent, and contextually grounded annotation across the dataset.

\subsection{Annotation Process}
\label{sec:sec_annotation_process}
The annotation workflow of \ficriticaled follows a structured, multi-stage process designed to ensure factual precision and consistency across annotators. Each annotator works within a controlled web-based interface displaying paired \texttt{HTML} text and its corresponding rendered page image. This dual-view setup lets annotators reference the visual layout when verifying line breaks, superscripts, or numerical formatting inconsistencies.

The annotation process begins with a \textbf{pre-screening phase}, where annotators review extracted HTML text for structural integrity and identify potential OCR or layout errors. Once validated, they proceed to the \textbf{entity marking phase}, where relevant critical fields are highlighted and enclosed within specialized span tags (e.g., \texttt{<Number>...</Number>}).

20\% of the data is annotated independently by at least two annotators to measure inter-annotator agreement. Upon completion, annotations are exported into JSON format, preserving the hierarchical document structure and linking each annotated entity to its source metadata. This structure enables downstream comparison against model-generated predictions for entity-level evaluation.

\begin{figure}[t]
    \centering
    \includegraphics[width=\columnwidth]{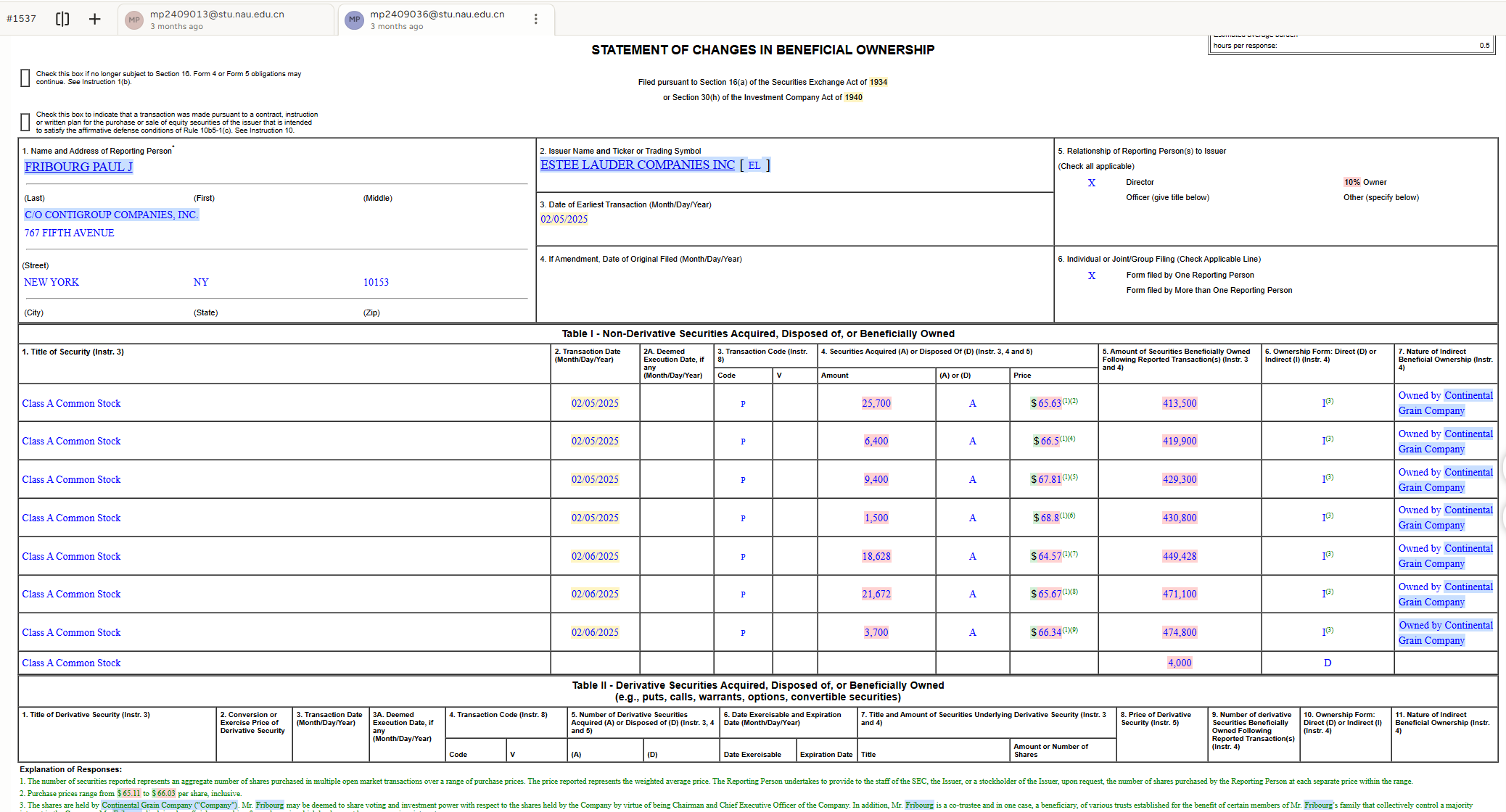}
    \caption{Annotation interface used in \ficriticaled. Annotators highlight entities directly within HTML while referencing rendered page images for layout validation.}
    \label{fig:annotation_ui}
\end{figure}

\subsection{Validation Guideline}
\label{sec:sec_validation_guideline}
To ensure reliability and reproducibility, \ficriticaled employs a multi-layered validation protocol combining automated checks, cross-annotator comparison, and expert review.

First, an \textbf{automated integrity validation} step scans all annotated files to detect malformed or nested span tags, missing entity attributes, or misaligned indices within the exported JSON schema.

Second, an \textbf{inter-annotator consistency check} evaluates pairwise agreement using token-level and entity-level overlap metrics. Entities with disagreement scores below a fixed confidence threshold are automatically queued for adjudication. Discrepancies are resolved through a consensus meeting facilitated by a lead annotator, who applies majority agreement rules and reviews edge cases (e.g., ambiguous table headers or multi-line numeric ranges).

Third, a \textbf{domain validation review} is conducted after every 100 annotated samples. In this phase, a senior financial expert randomly samples 5\% of the annotations to assess factual correctness. Feedback from this phase is incorporated into a continuously refined annotation guideline that codifies new patterns or exceptions encountered during labeling.

Finally, the validated annotations are version-controlled and re-exported to ensure traceability across dataset releases. The combination of automated validation, human consensus, and domain-level oversight establishes a high-confidence ground-truth benchmark for evaluating factual accuracy and numeric fidelity.

\subsection{Annotation Agreement Metric}
To assess the reliability and consistency of human annotation in \ficriticaled, we employ both \textbf{Cohen’s Kappa}~\cite{cohen1960coefficient} for pairwise annotator agreement and \textbf{Fleiss’ Kappa}~\cite{fleiss1971measuring} for overall multi-annotator reliability. These chance-corrected metrics provide a robust measure of agreement that accounts for the likelihood of random coincidence in categorical labeling.

Cohen’s Kappa ($\kappa$) quantifies the level of agreement between two annotators and is defined as:
\begin{equation}
\kappa = \frac{p_o - p_e}{1 - p_e},
\end{equation}
where $p_o$ represents the observed agreement (the proportion of items on which both annotators agree) and $p_e$ denotes the expected agreement based on random chance. A value of $\kappa = 1$ indicates perfect agreement, while $\kappa = 0$ reflects agreement equivalent to chance.

For more than two annotators, we adopt Fleiss’ Kappa, which generalizes the formulation of inter-rater agreement to multiple raters. It is computed as:
\begin{equation}
\kappa = \frac{\bar{P} - \bar{P}_e}{1 - \bar{P}_e},
\end{equation}
where $\bar{P}$ denotes the mean proportion of observed agreement across all annotation units, and $\bar{P}_e$ indicates the expected probability of agreement under a random labeling assumption. Similar to Cohen’s Kappa, higher values signify stronger consensus among annotators.

In \ficriticaled, we calculate Cohen’s Kappa for each annotator pair to evaluate pairwise consistency and Fleiss’ Kappa for the full group to measure overall reliability across five entity categories. Statistics in table \ref{tab:agreement} confirm a high level of agreement among the annotators, underscoring the robustness of the dataset’s annotation process.

\section{Deterministic-Rule-Guided LLM-As-Judge}
\label{sec:llm-as-judge-suppl}

\paragraph{Pipeline Overview.}
The evaluation proceeds in three stages:
\begin{enumerate}
  \item ~\textit{Entity Extraction}: \textit{BeautifulSoup} parses the gold HTML to enumerate all annotated entities by type, each paired with a $\pm$40-character context hint;
  \item ~\textit{LLM Judgment}: GPT-4o (\texttt{temperature=0}) receives the entity list, stripped gold text, and model prediction, and returns a JSON object indicating per entity whether it was \textit{found} and \textit{exactly matched};
  \item ~\textit{Deterministic Override}: any entity passing a normalized exact-match check is unconditionally set to $\text{correct}=\texttt{true}$, regardless of the LLM decision.
\end{enumerate}
\reviewhead{LLM-as-Judge prompt}
\begin{Verbatim}[
  fontsize=\scriptsize,
  frame=single,
  rulecolor=\color{black},
  framesep=3mm,
  labelposition=topline
]
JUDGE_PROMPT_TEMPLATE = """
#Instruction: You are an expert OCR results inspector for financial 
documents. You will be given:
1. A pre-extracted list of financially critical entities from the ground 
truth HTML. Each entity has: - "type": one of "Number", "Temporal", 
"Monetary Unit", "Reporting Entity", "Financial Concepts"
   - "value": the exact text of the entity
   - "context_hint": a short fragment of surrounding text that helps locate 
   it in the document
2. Ground truth HTML (entity tags already stripped, plain structure only) 
— for reference context.
3. Model-generated HTML that was produced from the same image.
Your task: for each entity in the pre-extracted list, determine whether 
the model-generated HTML contains that entity.
# Two-level judgment per entity
For each entity output TWO boolean fields:
- "found_in_prediction": true if the entity's value appears anywhere 
in the model HTML in a recognizable form — even if slightly corrupted 
(e.g., missing  comma, merged spaces, truncated).
  Set to false only if the entity is completely absent from 
  the prediction.
- "correct": true ONLY if the entity satisfies the strict 
matching rules below.
  (correct=true implies found_in_prediction=true; 
  found_in_prediction=true does NOT imply correct=true)
# PRIORITY RULE: exact match → always correct
If the matched_text (whitespace-normalized, case-insensitive) 
is identical to value, set correct=true.
The context_hint is only used to DISAMBIGUATE when the same 
value string appears in multiple different
locations in the prediction. It is NOT a veto: a different surrounding 
context does NOT make an otherwise exact match incorrect.
# Context-hint matching (apply loosely, only for disambiguation)
The context_hint is extracted from annotated HTML and may have minor 
formatting differences vs the prediction:
- Spaces may appear around currency symbols in the hint (e.g.,
"$ 66.59") but not in the prediction ("$66.59").
- Commas around numbers may differ in spacing (e.g., "per share, 
inclusive" vs "per share,inclusive").
- Apply whitespace-normalization and ignore spacing around $, 
%, (, ) when comparing context.
- The entity value itself (e.g., "66.59") must still appear in or 
very near the matched context region.

- IMPORTANT: Do NOT reject a match just because the context_hint 
has different spacing than the prediction.
  If the surrounding words match (e.g., "Purchase prices range 
\end{Verbatim}
\begin{Verbatim}[
  fontsize=\scriptsize,
  frame=single,
  rulecolor=\color{black},
  framesep=3mm,
  labelposition=topline
]
  from", "per share, inclusive"), that is
  sufficient context confirmation — minor punctuation/spacing 
  differences are expected and acceptable.
# Strict matching rules (apply when setting "correct")
- For "Number", "Temporal", "Monetary Unit":
  * correct=true if the exact numeric/date string (whitespace-normalized
  , case-insensitive) appears in
    the model HTML. If it appears multiple times, use context_hint 
    to pick the right occurrence.
  * Do NOT mark correct if the string is corrupted, truncated, or 
  merged with adjacent characters.
    Examples of NOT correct: "ERIBOURGPAUL" for "FRIBOURG PAUL J"
    ; "25700" for "25,700"; "0.5" for "0.50".
  * Mark found_in_prediction=true if a close but not exact version 
  appears (e.g., "25700" for "25,700").
- For "Reporting Entity", "Financial Concepts":
  * correct=true if the full string appears (whitespace-normalized
  , case-insensitive) in the model HTML.
    If it appears multiple times, use context_hint to pick the 
    right occurrence.
  * Minor run-together spaces within a single token (e.g., 
  "UNITEDSTATES" for "UNITED STATES") may be
    marked correct if still fully recognizable — record actual 
    text in matched_text.
  * Missing words, swapped words, or substituted characters → 
  correct=false, but may be found_in_prediction=true.
# Step 1: Match entities
For each entity in the pre-extracted list, output a diagnostic record 
with:
  - "type", "value", "context_hint" (copied from input)
  - "found_in_prediction": true/false (recognizable attempt exists 
  in prediction)
  - "correct": true/false (exact match per strict rules above; exact 
  match always overrides context check)
  - "matched_text": the actual text found in the model HTML (empty 
  string "" if not found at all)
# Step 2: Compute counts
The total_entities counts come directly from the pre-extracted list 
— do NOT re-count from the ground truth HTML.
Compute:
  - found_entities = number of diagnostics where found_in_prediction 
  = true
  - correct_entities = number of diagnostics where correct 
  = true
  - found_entities_with_X_type and correct_entities_with_X_type 
  for each of the 5 types
  - entity_accuracy = correct_entities / found_entities * 100, 
  rounded to 2 decimal places (0.0 if found_entities = 0)
# Step 3: Output
Output exactly one valid JSON object:
{
  "total_entities": <int>,
  "total_entities_with_Number_type": <int>,
  "total_entities_with_Temporal_type": <int>,
  "total_entities_with_Monetary_Unit_type": <int>,
  "total_entities_with_Reporting_Entity_type": <int>,
  "total_entities_with_Financial_Concepts_type": <int>,
  "found_entities": <int>,
  "found_entities_with_Number_type": <int>,
  "found_entities_with_Temporal_type": <int>,
  "found_entities_with_Monetary_Unit_type": <int>,
  "found_entities_with_Reporting_Entity_type": <int>,
  "found_entities_with_Financial_Concepts_type": <int>,
  "correct_entities": <int>,
  "correct_entities_with_Number_type": <int>,
  "correct_entities_with_Temporal_type": <int>,
  "correct_entities_with_Monetary_Unit_type": <int>,
  "correct_entities_with_Reporting_Entity_type": <int>,
  "correct_entities_with_Financial_Concepts_type": <int>,
  "entity_accuracy": <float 0-100>,
  "overall_explanation": "<1-2 sentences summarizing model quality>",
  "entity_diagnostics": [
    {"type": "...", "value": "...", "context_hint": "...",
     "found_in_prediction": true, "correct": false, "matched_text": "25700"},
    ...
  ]
}
Rules:
* Output only valid JSON (no extra text).
* All count fields must be integers, not strings.
* entity_accuracy must be a float (e.g. 85.71), NOT a string like "85.71%".
* entity_accuracy = 
correct_entities / found_entities * 100 (NOT correct / total).
\end{Verbatim}
\begin{Verbatim}[
  fontsize=\scriptsize,
  frame=single,
  rulecolor=\color{black},
  framesep=3mm,
  labelposition=topline
]
* total_entities counts MUST match the counts from the pre-extracted list.
* REMINDER: if matched_text equals value (normalized), correct MUST be true — 
do not set correct=false just because the surrounding context in the 
prediction differs from context_hint.
"""
\end{Verbatim}

\reviewhead {LLM-as-Judge output sample}
\begin{Verbatim}[
  fontsize=\scriptsize,
  frame=single,
  rulecolor=\color{black},
  framesep=3mm,
  labelposition=topline
]
{'total_entities': 65,
 'total_entities_with_Number_type': 37,
 'total_entities_with_Temporal_type': 10,
 'total_entities_with_Monetary_Unit_type': 7,
 'total_entities_with_Reporting_Entity_type': 11,
 'total_entities_with_Financial_Concepts_type': 0,
 'found_entities': 64,
 'found_entities_with_Number_type': 36,
 'found_entities_with_Temporal_type': 10,
 'found_entities_with_Monetary_Unit_type': 7,
 'found_entities_with_Reporting_Entity_type': 11,
 'found_entities_with_Financial_Concepts_type': 0,
 'correct_entities': 64,
 'correct_entities_with_Number_type': 36,
 'correct_entities_with_Temporal_type': 10,
 'correct_entities_with_Monetary_Unit_type': 7,
 'correct_entities_with_Reporting_Entity_type': 11,
 'correct_entities_with_Financial_Concepts_type': 0,
 'entity_accuracy': 98.36,
 'overall_explanation': 'The model successfully identified and correctly
 matched nearly all entities, with only one minor discrepancy in a number 
 format.',
 'entity_diagnostics': [{'type': 'Number',
   'value': '10%',
   'context_hint': 'X Director 10% Owner Officer (give title below) Other',
   'found_in_prediction': True,
   'correct': True,
...
 'not_found_entities_with_Number_type': 1,
 'not_found_entities_with_Temporal_type': 0,
 'not_found_entities_with_Monetary_Unit_type': 0,
 'not_found_entities_with_Reporting_Entity_type': 0,
 'not_found_entities_with_Financial_Concepts_type': 0}
\end{Verbatim}

\paragraph{Design Details.}
\textit{Context Window.}
Many financial documents repeat identical values across different line items (e.g., the number ``1,200'' may appear in both revenue and headcount rows of the same table). To prevent the judge from conflating these, each entity is accompanied by the 40 characters immediately preceding and following it in the gold text. For instance, the entity \texttt{1,200} might be disambiguated by the hint \texttt{``...total revenue (in millions) \underline{1,200} for the fiscal year...''}, directing the judge to verify that specific occurrence rather than any surface match.

\textit{Normalization and Deterministic Override.}
Before applying the override, both the predicted and gold entity values are normalized: lowercased, whitespace-collapsed, and stripped of punctuation (commas, currency symbols, trailing periods). If the normalized strings are identical, the entity is forced correct. This guards against LLM false negatives on unambiguous cases.

\textit{Found vs.\ Correct.}
An entity is \textit{found} if the judge locates it (or a semantically equivalent expression) anywhere in the prediction; it is \textit{correct} if the reproduced value is additionally judged to be factually exact. An entity may be found but incorrect (e.g., a number located in the wrong context), or neither found nor correct (entirely absent from the prediction).

\begin{figure}[t]
  \centering

  \begin{minipage}{\columnwidth}
    \centering
    \includegraphics[width=\linewidth]{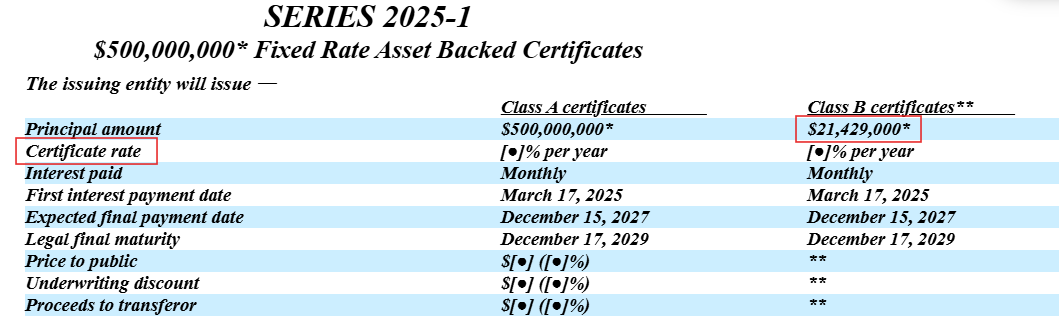}
    
    \small (a) Part of financial statement investment plan explained
  \end{minipage}

  \vspace{0.5em}

  \begin{minipage}{\columnwidth}
    \centering
    \includegraphics[width=\linewidth]{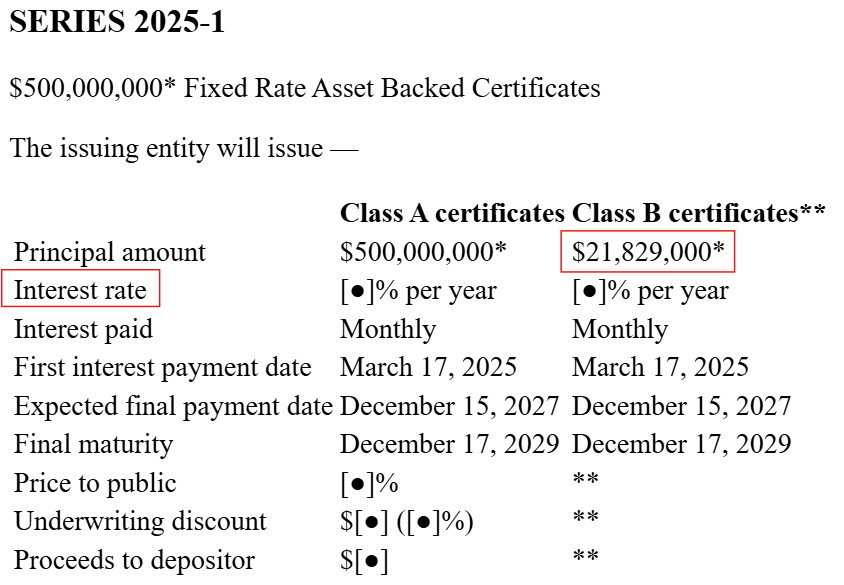}
    
    \small (b) Corresponding LLM output in HTML
  \end{minipage}

  \caption{Human alignment with evaluation results on one of the GPT-4o outputs. Highlighted areas indicate where OCR results on critical fields are incorrect.}
  \label{fig:alignment_case}
\end{figure}

\paragraph{Human Alignment Validation.}
To validate the automated judge, 15 outputs from Paddleocrv5, gpt-4o, and gpt-5 each (45 total) were independently evaluated by a human expert using the same found/correct schema. We adopt \textit{Document-Level Strict Agreement}: judge and human must agree on every entity within a document to count as aligned; a single divergence constitutes disagreement. GPT-4o achieved \textbf{95\%} agreement after iterative prompt and evaluation workflow refinement. Figure ~\ref{fig:alignment_case} shows a representative alignment example:

\paragraph{Case: Overall FFA = 90\%}
On Figure ~\ref{fig:alignment_case}, both human experts and the LLM-as-Judge classify the output as factually unreliable.  The LLM-as-Judge hallucinates the row name "Certificate rate" to "Interest rate," as it associate with the "Interest paid" row name below. The model also misread the monetary amount.
\textit{LLM output:}
\begin{Verbatim}[
  fontsize=\scriptsize,
  frame=single,
  rulecolor=\color{black},
  framesep=3mm,
  labelposition=topline
]
{
    "total_entities": 23,
    ...
    "found_entities_with_Number_type": 3,
    ...
    "found_entities_with_Financial_Concepts_type": 10,
    ...
    "correct_entities_with_Number_type": 2,
    ...
    "correct_entities_with_Financial_Concepts_type": 9,
    "entity_accuracy": 90.0,
    "overall_explanation": "The model accurately identified most entities, 
    especially reporting entities and financial concepts, but struggled 
    with exact numeric matches."
    }
\end{Verbatim}

\textit{Human expert explanation:}

\begin{Verbatim}[
  fontsize=\scriptsize,
  frame=single,
  rulecolor=\color{black},
  framesep=3mm,
  labelposition=topline
]
Total entities: 23
Numeric entities: 3
...
Financial Concepts: 10
...
Wrong numeric entities: 1
...
Wrong reporting entities: 1
Overall:  The model performs well overall, except errors due to halluciantion 
on "Interest rate" and wrong numerical entity of "21,829,000 
(should be "21,429,000").
\end{Verbatim}
These observations correspond precisely to the errors and overall judgment from GPT-4o output. 

\section{Model Failure Analysis}
\label{sec:error_cases}
This section analyzes model failures beyond overall FFA scores. We first describe the error-analysis setup, then examine two notable aggregate patterns: elevated missing-output behavior in some proprietary models and GLM-OCR's strong performance on financial concepts. We finally present representative error cases illustrating how different failure modes appear in practice.

\subsection{Error Analysis Setup}
\label{sec:error_analysis_setup}

We perform error analysis by comparing annotated ground-truth content with model-predicted OCR outputs after both are rendered in HTML. Specifically, the ground-truth document and the corresponding model output are loaded into Label Studio, where annotators inspect them side by side and identify financially meaningful discrepancies. During review, they assess whether a model preserves both local transcription and the financial meaning of each critical field under its rendered context, rather than only surface-level textual overlap.

We organize reviewed examples along two dimensions: \textbf{document modality} and \textbf{document complexity}. We first categorize documents into three modality types:

\begin{itemize}
    \item \textbf{Table-only}: primarily tabular content, including table of contents pages or documents containing only tables, with no surrounding text beyond titles or headers.
    \item \textbf{Text-only}: contain only text without tables, including one or more paragraphs, with or without section headings.
    \item \textbf{Mixed modality}: contain both tables and text, where the text extends beyond titles, headers, or annotations.
\end{itemize}

Within each modality, we further divide examples into low, medium, and high complexity using modality-specific criteria:

\begin{itemize}
    \item \textbf{Table-only}: low = simple tables such as table of contents pages; medium = clearly structured financial tables organized in a straightforward row-by-row format; high = complex or irregular financial tables, such as tax forms, where information cannot be easily interpreted line by line.
    \item \textbf{Text-only}: low = plain text without section headings or numeric financial indicators such as numbers, dates, or monetary values; medium = text with either section headings or numeric indicators such as numbers, dates, or monetary values; high = text with both section headings and numeric financial indicators.
    \item \textbf{Mixed modality}: low = documents containing tables with simple accompanying text without section headings or additional numeric indicators; medium = documents containing tables and text with either section headings or numeric indicators; high = documents containing tables and text with both section headings and numeric indicators.
\end{itemize}

Using this setup, we identify representative error cases and map them to the broader qualitative taxonomy introduced in Table~\ref{tab:error_taxonomy}.

\subsection{Missing-Output Effects in Proprietary Models}
GPT-4o and GPT-5 exhibit unusually low FFA partly because they have much higher missing-output rates than most other models in our evaluation. Their overall missing rates are 23.12\% and 13.41\%, respectively, compared with 1.94\% for Qwen3-VL-8B-Instruct, 1.76\% for Llama-4-Maverick, and 1.73\% for Claude-Sonnet-4.6. The gap is especially large for numeric fields, where the missing rates rise to 31.35\% for GPT-4o and 24.0\% for GPT-5. In manual inspection, many failures appear as incomplete or prematurely cut-off HTML rather than straightforward recognition errors. Because this pattern persists even after substantially increasing the output budget, we do not attribute it mainly to length limits. Instead, it may reflect more conservative generation behavior on long, structured, and potentially sensitive financial pages, which is consistent with OpenAI's description of GPT-5 safe-completion and stronger caution under ambiguous-risk settings, while also suggesting a model tendency to abstain under difficult document conditions ~\cite{hurst2024gpt,openai_gpt5}.

\subsection{Why Does GLM-OCR Reach 100\% FC-FFA?}
\paragraph{GLM-OCR on financial concepts.}
GLM-OCR achieves \textbf{100\% FC-FFA} in our evaluation, indicating exceptional extraction of financial-concept spans such as line items and table labels. Because financial concepts are typically expressed as plain-text semantic units rather than symbol-sensitive values, performance on this field type depends heavily on robust text recognition and structure-aware parsing. This result is likely related to GLM-OCR's OCR-specialized design: a compact encoder--decoder architecture with a dedicated visual encoder, a GLM decoder, and a two-stage pipeline that first performs layout analysis and then applies region-level recognition~\cite{duan2026glmocrtechnicalreport}. This design is particularly well aligned with financial-concept extraction, where accurate isolation of row labels and preservation of local document structure are critical. More broadly, the result suggests that OCR-specialized document models can hold a clear advantage over general-purpose multimodal models on plain-text financial semantics, especially when success depends more on structural fidelity than symbolic precision.

\subsection{Representative Model Error Cases}
In Table~\ref{tab:error_taxonomy}, we presents the error taxonomy and the representative models and model failure examples:
\begin{table*}[t]
\setlength{\abovecaptionskip}{3pt}
\setlength{\belowcaptionskip}{0pt}
\centering
\small
\caption{Qualitative taxonomy of observed Model OCR failure patterns. OCR-specialized systems tend to show fixed and repetitive recognition errors, while general multimodal LLMs exhibit more variable and generative failure modes.}
\label{tab:error_taxonomy}
\resizebox{\linewidth}{!}{
\begin{tabular}{p{2.0cm}p{4.5cm}p{3.0cm}p{6.0cm}}
\toprule
\setlength{\abovecaptionskip}{3pt}
\textbf{Error Type} & \textbf{Description} & \textbf{Common Models} & \textbf{Example} \\
\midrule
Header misalignment & Correct value is transcribed but attached to the wrong column or local header context. & Llama-family& "6. Ownership Form: Direct (D) or Indirect (I) (Instr. 4)" value aligned to "7. Nature of Indirect Beneficial Ownership (Instr. 4)". \\
Heading recognition error & The model fails to correctly recognize section titles or subtitles, either by omitting them or transcribing them incorrectly, leading to incorrect document structure or emphasis. & All with different levels of severity & gemma-3n-E4B-it failed to recognize section subtitle "Credit card rates may decline without a corresponding change in the amounts needed to pay the certificates, which could result in a delay or reduction in payments of your certificates." \\
Hallucinated continuation & Similar sentence prefix causes the model to copy the continuation of a previously seen sentence instead of the correct local text. & Gemma-3n-E4B-it, GPT-4o & Document started to repeating the sentence "the undergaining and account management standards of card issuers and could restrict the ability of AEN’s of its affiliates to take" on the last paragraph. \\
Fixed lexical corruption & The same word or token is repeatedly mistranscribed across pages. & DeepSeek-OCR, MinerU2.5  &  "transferor" was changed into "transfessor", "transforor", "transferror", or "transfeder" etc. across different documents. \\ 
Format-sensitive spacing & OCR errors consistently occur around dashes, punctuation, or spacing-sensitive formats. & DeepSeek-OCR, MinerU2.5 & "Certain Legal Aspects of the Receivables — Consumer Financial Products Regulation" was changed into "Certain Legal Aspects of the Receivables - Consumer Financial Products Regulation" \\
Sentence-local corruption & An OCR error confined to a single sentence, where the model omits, inserts, rewrites, or distorts content within that sentence without affecting other parts of the document. & Llama-family &  Document missed "and the collateral securing the certificates" \\
Unstable free-form corruption & Errors vary substantially across samples, even under similar layouts or textual contexts. & Gemma-3n-E4B-it, GPT-4.o, GPT-5 & GPT-5 missed the whole paragraph that even including financial facts.\\
\bottomrule
\end{tabular}}

\end{table*}

\section{Limitations}
While \ficriticaled advances fact-level evaluation for financial OCR, several limitations remain.

First, \ficriticaled mainly covers U.S. financial documents with relatively standardized visual conventions; broader international, multilingual, and handwritten settings may introduce challenges not captured here.

Second, our annotation pipeline uses rendered HTML as the structural reference. Although stable and reproducible, it may differ from native PDF or scanned-document artifacts in real OCR deployments.

Third, \ficriticaled evaluates factual correctness at the entity level, but does not yet capture cross-page linking, hierarchical financial relationships, or long-context numerical grounding across document collections.

These limitations suggest future work in multilingual, hybrid human--machine evaluation for edge cases, and richer benchmarks combining OCR, layout understanding, and financial reasoning.

\section{Potential Risks and Misuse}
\ficriticaled is intended to advance high-precision financial OCR research, but several risks should be noted.

First, although the benchmark uses publicly available financial documents, stronger OCR may enable more effective extraction of sensitive information from documents not intended for automated analysis. Responsible use requires compliance with privacy, regulatory, and data-handling requirements.

Second, the LLM-as-Judge framework is designed only for evaluation and should not be treated as an authoritative tool for auditing, compliance, or legally binding document verification.

Third, like any dataset, \ficriticaled may reflect geographic, regulatory, and formatting biases in its source documents. Models trained or tuned only on this dataset may overfit to these conventions and generalize poorly elsewhere.

Finally, highly accurate OCR may enable large-scale financial data extraction for questionable uses, including unauthorized scraping, adversarial market strategies, or amplification of misleading financial narratives. Researchers and practitioners should consider both beneficial and harmful downstream impacts.

\section{Ethical Considerations and Licensing}
All documents come from publicly available financial filings released under open-access or research licenses. No proprietary or confidential information is included. The dataset is intended for research on document understanding and factual accuracy; commercial use requires separate compliance review. The benchmark complies with ACM data ethics policies.


\end{document}